\documentclass{article} 
\usepackage[preprint, nonatbib]{neurips_2020}
\usepackage{times}

\pdfoutput=1

\usepackage{amsmath,amsfonts,bm}

\def\eqref#1{equation~\ref{#1}}

\def\1{\bm{1}}

\DeclareMathAlphabet{\mathsfit}{\encodingdefault}{\sfdefault}{m}{sl}
\SetMathAlphabet{\mathsfit}{bold}{\encodingdefault}{\sfdefault}{bx}{n}

\usepackage{hyperref}
\usepackage{url}
\usepackage{enumerate}
\usepackage{graphicx,subcaption}
\usepackage{bbold}
\usepackage{float}
\usepackage[ruled,vlined]{algorithm2e}
\usepackage{booktabs}
\usepackage{subcaption}

\usepackage{multirow}
\usepackage{color}
\usepackage{wrapfig}
\usepackage{relsize}
\usepackage{siunitx}
\usepackage{microtype}
\title{A Study on Efficiency in Continual Learning \\ Inspired by Human Learning}

\author{Philip J. Ball \thanks{Work done as part of an internship at Microsoft Research. Correspondence \texttt{ball@robots.ox.ac.uk}} \\
Machine Learning Research Group \\
University of Oxford\\
\And
Yingzhen Li\\
Microsoft Research Cambridge \\
\AND
Angus Lamb\\
Microsoft Research Cambridge \\
\And
Cheng Zhang \\
Microsoft Research Cambridge \\
}

\begin{document}

\maketitle

\begin{abstract}
Humans are efficient continual learning systems; we continually learn new skills from birth with finite cells and resources. Our learning is highly optimized both in terms of capacity and time while not suffering from catastrophic forgetting. In this work we study the efficiency of continual learning systems, taking inspiration from human learning. In particular, inspired by the mechanisms of sleep, we evaluate popular pruning-based continual learning algorithms, using PackNet as a case study. First, we identify that weight freezing, which is used in continual learning without biological justification, can result in over $2\times$ as many weights being used for a given level of performance. Secondly, we note the similarity in human day and night time behaviors to the training and pruning phases respectively of PackNet. We study a setting where the pruning phase is given a time budget, and identify connections between iterative pruning and multiple sleep cycles in humans. We show there exists an optimal choice of iteration v.s. epochs given different tasks.
\end{abstract}
\section{Introduction}

Humans are continual learning systems that have been very successful at adapting to new situations while not forgetting about their past experiences \cite{lieberman_2011}. In many real-life machine learning applications, continual learning is also required. For example, a model trading on the financial markets makes decisions during the day, then trains on new market data overnight. A successful model should adapt to these new distributions while maintaining good performance on the past data. Furthermore, for practical use the model requires the following two desirable properties. Firstly, the model should be memory efficient and operate within a hardware budget. Secondly, the model should be time efficient in two aspects: (i) data which arrives periodically needs to be processed as a brand new task; (ii) the training time for the model cannot exceed some predefined time budget (i.e., overnight training). Humans have evolved to handle continual learning tasks efficiently while satisfying both desiderata. For example, humans operate with finite number of neurons \cite{sorrells_human_2018}, analogous to a machine learning model with a prefixed (and potentially limited) memory budget. Humans also receive information from the environment during the day, and sleep for 7-9 hours to consolidate the information in the evening, similar to training a machine learning model with limited time budget.

Continual learning methods based on weight freezing \cite{prognets2016, packnet2018, piggyback2018, compactingpicking2019} allocate specific parts of the neural network to different tasks, thereby addressing catastrophic forgetting by construction \cite{forgetting1989, forget1999}. As the model faces a limited memory budget, pruning needs to be performed in order to free-up parts of the network for future tasks.
This is analogous to human learning, where learning new tasks can be viewed as perception during the day, while pruning corresponds to sleep during the night. Specifically, during non-rapid eye movement (NREM) sleep, weak synapses are removed \cite{humanpruning2006}.
As humans typically experience multiple NREM/REM sleeping cycles, this inspires the application of iterative pruning, a technique that has been studied in supervised learning \cite{pruning1989,braindamage1990,iterative2015}, to the continual learning setting.


In this work, we analyze a popular pruning based continual learning algorithm, PackNet \cite{packnet2018}, through the lens of human learning; this framing naturally lends itself to a more practical viewpoint, and helps us identify issues that may prevent their application to real-world systems. In particular, we would like to answer two questions. Firstly, does weight freezing in continual learning, which has no biological justification, limit memory efficiency (Section \ref{sec:weight})? Secondly, if we apply iterative pruning to PackNet, what is the optimal trade-off between iteration count and re-training epochs in pruning-based continual learning methods, given a limited time budget (Section \ref{sec:budget})?

\section{Weight Freezing is Memory Inefficient}
\label{sec:weight}

Neuroplasticity is key for human learning in new environments and situations, where it manifests during sleep (even in adults \cite{adultneuroplastic2014, zhang2017neural}). 
This is in contrast to weight freezing techniques in continual learning where, once a task is learned, the weights dedicated to those tasks are then frozen for the rest of training. We hypothesize that a degree of plasticity must be retained in order to ensure continued adaptability, and that weight freezing reduces the ability of the network to learn new tasks.

\vspace{-0.7em}
\paragraph{Memory Efficiency}
We define memory efficiency as the number of weights required for a specific level of performance on a continual learning task. For example, if network A achieves $99\%$ average accuracy over all tasks and has $60\%$ of its weights still free for future tasks, while network B achieves the same accuracy but has $90\%$ of its weights free, we calculate the additional weight compression of B as follows: $\frac{1 - 0.60}{1 - 0.90}=4\times$. Therefore we say that network B is $4\times$ more memory efficient than A.

We demonstrate the memory inefficiency of weight freezing by empirically comparing PackNet \cite{packnet2018} against two idealized scenarios: a) multitask training; b) perfect memory with no weight-freezing.
\begin{itemize}
\vspace{-0.3em}
    \item 
\textbf{Multitask Method (Multinet):} We have access to all the data at once, and can therefore learn a parameterization that optimizes the solution across all tasks simultaneously.
    \item 
\textbf{Full Rehearsal Memory Method (Memory):} We store all seen data in a memory buffer, and at each task, train on both the new data and all the previously seen data. We do not freeze the weights, instead we expand all weight masks to accommodate these new tasks.
\end{itemize}

\subsection{Experiments and Results}
\label{sec:weightefficiencyresults}
In reality, the total number of tasks is often unknown, and we therefore should aim to fit as many tasks as possible with a single network. In this case, we define a final pruning amount as $p$ for PackNet, which means that fraction $p$ of the weights will be available to learn future tasks. For a fair comparison, we train all approaches with the same time budget.

We test on both binarized versions of MNIST \cite{splitmnist2017} and CIFAR10. For the former we use a 2 hidden layer MLP, and for the latter we use a simplified VGG16 architecture \cite{vgg15}. See Appendices \ref{app:imp} and \ref{app:match} for implementation and further experiment details respectively.

\begin{figure}[t]
\centering
\begin{subfigure}{.45\textwidth}
    \centering
    \includegraphics[width=\textwidth]{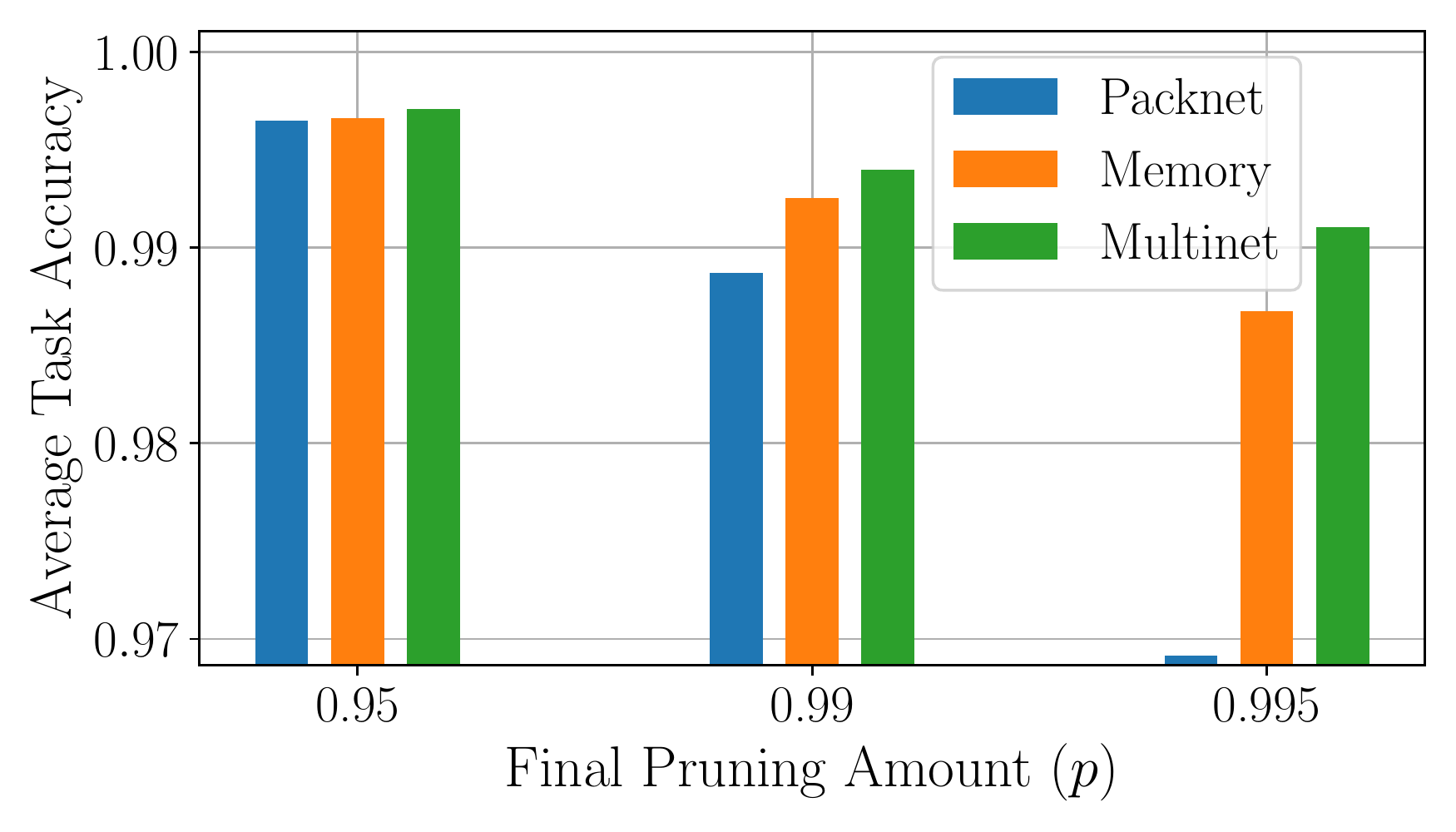}
    \label{fig:compare_all_mnist}
\end{subfigure}
\hspace{0.5cm}
\begin{subfigure}{.45\textwidth}
    \centering
    \includegraphics[width=\textwidth]{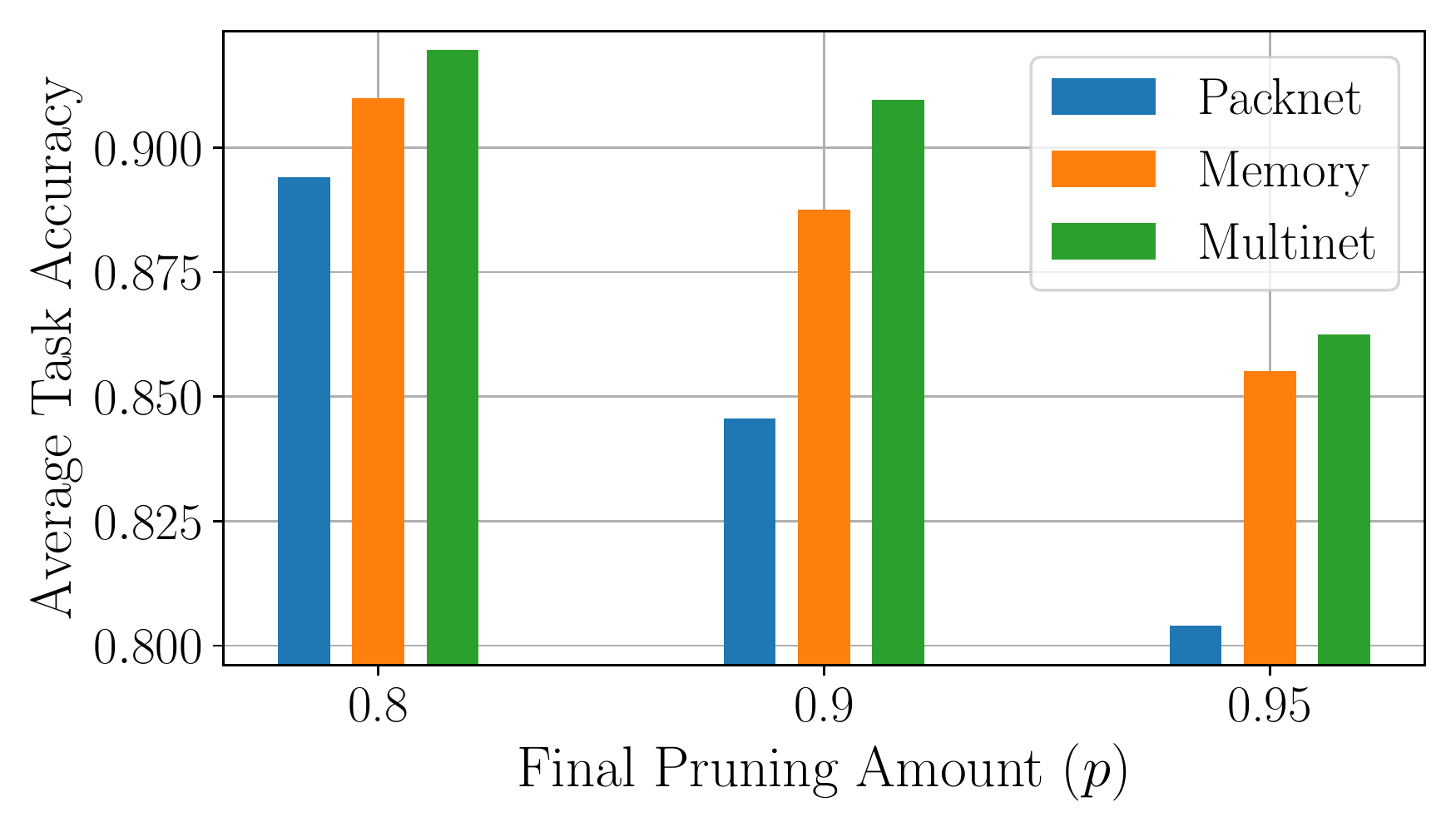}
    \label{fig:compare_all_cifar10}
\end{subfigure}
\vspace{-0.5cm}
\caption{PackNet v.s. Idealized Cases. \textbf{Left:} MNIST. \textbf{Right:} CIFAR10.}
\label{fig:compare_all}
\end{figure}

Figure \ref{fig:compare_all} shows the accuracy results of each approach for a given final pruning amount $p$.
We observe that PackNet performs the worst, and that the disparity in performance increases with higher pruning amount $p$. Table \ref{tab:matching} shows the results of a matching experiment (details in Appendix \ref{app:match}), where we match the performance of a PackNet network at a given amount of pruning with the other two approaches, and report the corresponding compression ratios. We observe that as we increase pruning, the relative memory efficiency of PackNet worsens compared to the other approaches. For MNIST, this means PackNet can use up to $3.32\times$ more weights to achieve the same performance as the multitask approach. In the case of CIFAR10, the memory and multitask approaches do similarly well, usually being over $2\times$ more memory efficient than PackNet. This suggests that, regardless of neural network architecture, weight freezing is not memory efficient.

\vspace{-0.4cm}
\begin{table}[t]
    \centering
    \small
    \begin{subtable}{.5\textwidth}
    \centering
    \begin{tabular}{@{}ccc@{}}
    \toprule
    \multirow{2}*{\vspace{-0.1cm} PackNet Pruning ($p$)} & \multicolumn{2}{c}{Compression} \\ \cmidrule{2-3}
     & Memory & Multinet  \\ \midrule
        $95.0\%$ & $1.17\times$ & $1.63\times$ \\
        $99.0\%$ & $1.45\times$ & $2.77\times$ \\
        $99.5\%$ & $1.54\times$ & $3.32\times$ \\
    \bottomrule
    \end{tabular}
    \end{subtable}%
    \begin{subtable}{.5\textwidth}
    \centering
    \begin{tabular}{@{}ccc@{}}
    \toprule
    \multirow{2}*{\vspace{-0.1cm} PackNet Pruning ($p$)} & \multicolumn{2}{c}{Compression} \\ \cmidrule{2-3}
     & Memory & Multinet  \\ \midrule
        $80.0\%$ & $1.60\times$ & $2.82\times$ \\
        $90.0\%$ & $2.25\times$ & $2.26\times$ \\
        $95.0\%$ & $2.08\times$ & $2.27\times$ \\
    \bottomrule
    \end{tabular}
    \end{subtable}%
    \vspace{0.3cm}
    \caption{PackNet v.s. Idealized Cases, Matching Performance. \textbf{Left:} MNIST. \textbf{Right:} CIFAR10.}
    \label{tab:matching}
\end{table}
\section{Iterative Pruning on a Time Budget}
\label{sec:budget}
Sleeping in humans is beneficial for memory because it aids with experience consolidation \cite{rasch_about_2013}. This has analogies to the pruning phase in some continual learning methods; having experienced (i.e., trained) during the day, we then consolidate (i.e., prune and retrain) during the night. Humans have sleep cycles \cite{cycle1968}, where oscillation between REM and NREM phases occurs. This is analogous to `iterative pruning', where retraining can be viewed as the REM phase, and pruning as the NREM phase, where weak weights/synapses are removed.



We wish to understand if the benefit of multiple dream cycles in human sleep translates to continual learning with iterative pruning. However, previous studies do not consider time constraints for the pruning phase \cite{iterative2015, Renda2020Comparing}, which differs from human sleep that lasts for a fixed amount of time. Indeed it is crucial to consider the time budget in many application scenarios, such as the aforementioned overnight training of a financial trading model. Therefore the prune-phase time budget should be fixed for iterative pruning in order to enable applications to realistic continual learning tasks.

\begin{wrapfigure}[17]{r}{0.45\textwidth}
 \begin{minipage}{0.5\textwidth}
 \vspace{-14pt}
 \scalebox{0.9}{
\begin{algorithm}[H]
\SetAlgoLined
\SetKwInOut{Input}{Input}
\SetKwInOut{Init}{Initialize}
 \Input{Number of iterative pruning rounds $N$, Tasks $t_i \in T$, Final prune proportion $p$}
 \Init{Initial neural network $M$ (having weights $M_W$), tasks so far $i=1$, initial mask $m_0 = \mathbb{1}_{|M_W|}$}
 Calculate $z$ using Algorithm \ref{alg:getprunepct}\;
 \While{$i < |T|$}{
  Train $M$ on task $t_i$, freezing weights $M_W \odot m_{i-1}$\;
  \For{n = 1, \dots, N}{
    Prune network by $z\%$\;
    Retrain network on task $t_i$\;}
  Save weight mask $m_i$\;
  $i = i+1$\;
 }
 \caption{PackNet with Iterative Pruning}
 \label{alg:packnetiterative}
\end{algorithm}
}
\end{minipage}
\end{wrapfigure}
\vspace{-0.5em}
\paragraph{Iterative Pruning in Continual Learning}
Compared to `one-shot' pruning, iterative pruning successively prunes and retrains a network until a desired amount of weight reduction. For example, if we want $50\%$ weight reduction, we can take a one-shot approach and prune $50\%$ once then retrain, or we can iteratively prune twice, each time pruning $25\%$ of the total weights and retraining.
PackNet uses one-shot pruning for network weight reduction; we extend this algorithm to use iterative pruning and present this in Algorithm \ref{alg:packnetiterative}. If $N=1$, we retrieve the original PackNet algorithm.

\vspace{-0.5em}
\paragraph{Computational Efficiency}
Iterative pruning in its original form runs for $E + (R \times N)$ total epochs \cite{pruning1989}, where $E$ is epochs dedicated to training, $R$ is the epochs per phase of retraining, and $N$ is the rounds of iterative pruning, therefore our computational requirements scale as $\mathcal{O}(N)$.
In experiments we fix the time budget for pruning to $B = R \times N$ and vary the number of pruning rounds $N$.

\subsection{Experiments and Results}

\begin{figure}[t]
    \centering
    \begin{subfigure}{.32\textwidth}
    \includegraphics[width=\textwidth]{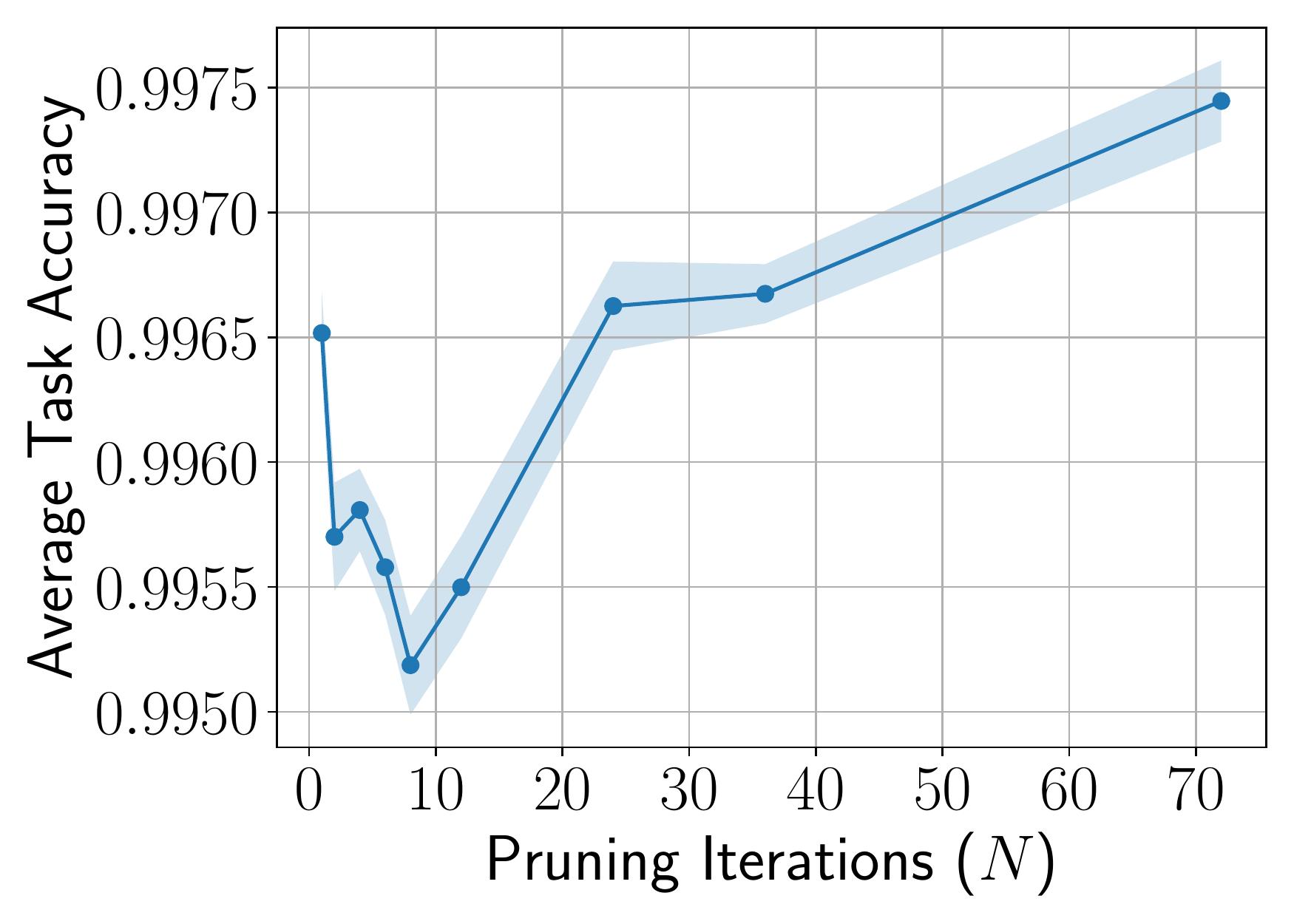}
    \caption{PackNet: 0.95 Pruning}
    \label{fig:packnet095mnist}
    \end{subfigure}%
    \begin{subfigure}{.32\textwidth}
    \includegraphics[width=\textwidth]{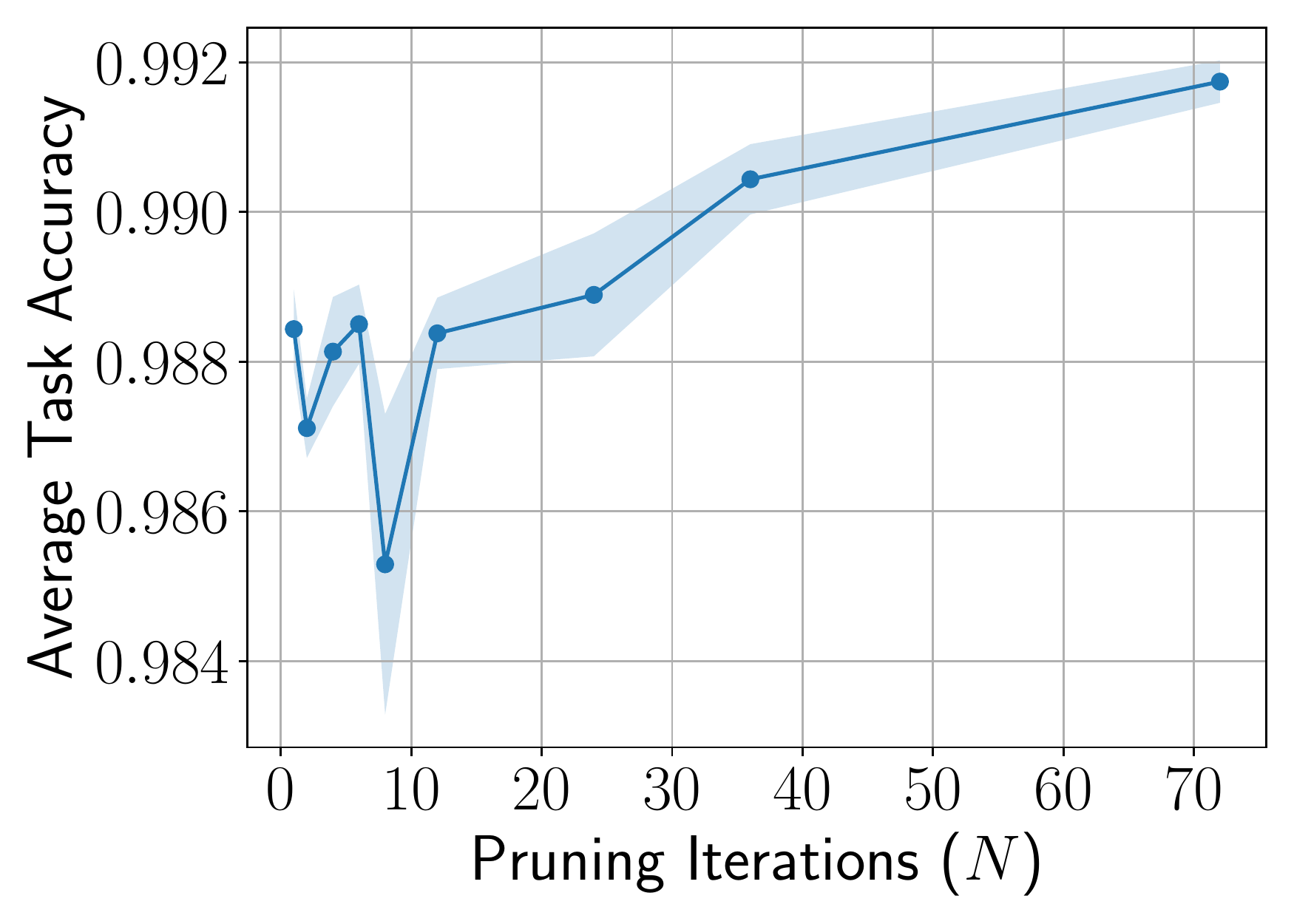}
    \caption{PackNet: 0.99 Pruning}
    \label{fig:packnet099mnist}
    \end{subfigure}%
    \begin{subfigure}{.32\textwidth}
    \includegraphics[width=\textwidth]{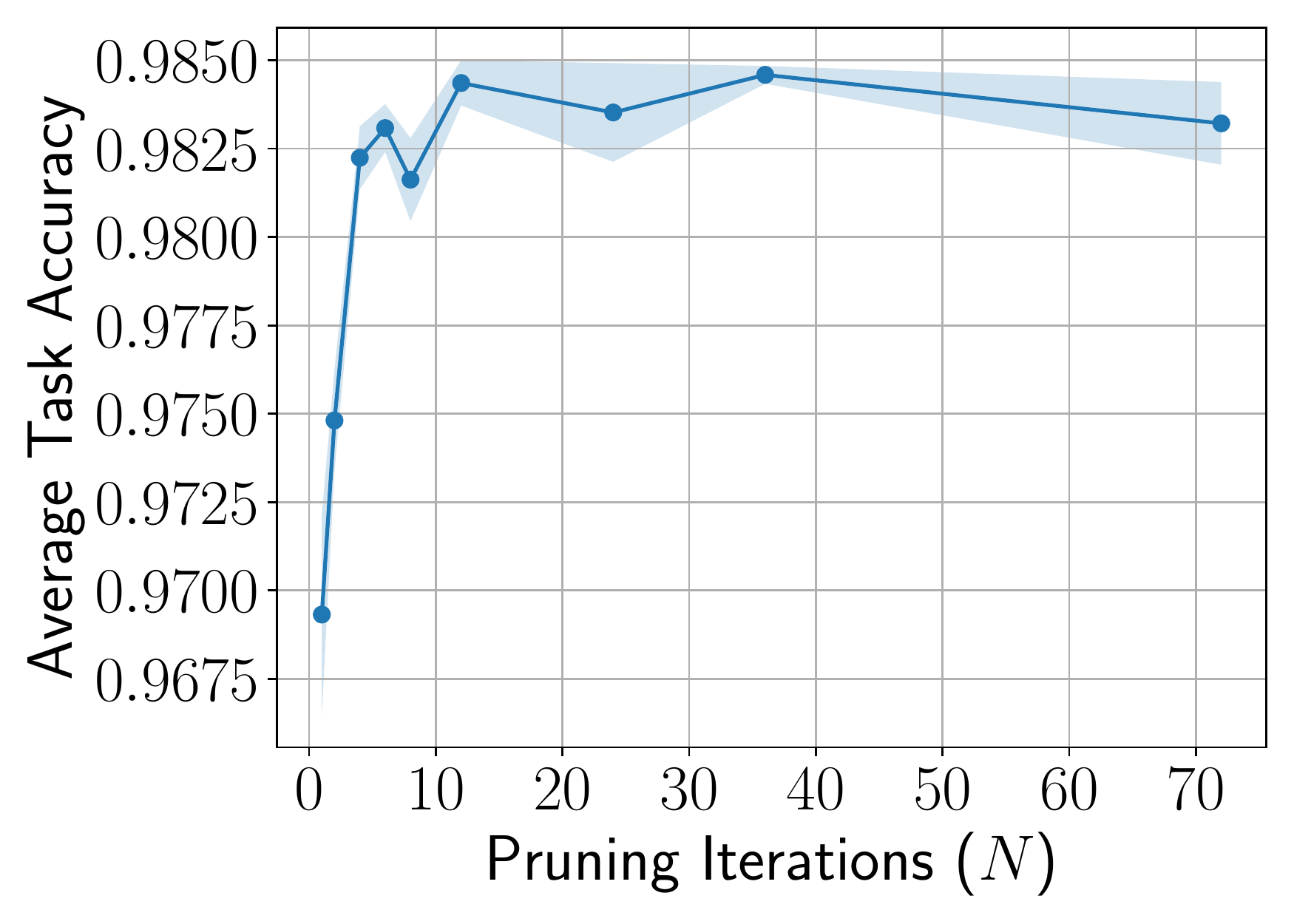}
    \caption{PackNet: 0.995 Pruning}
    \label{fig:packnet0995mnist}
    \end{subfigure}%
    \label{fig:iter_mnist}
    \\
    \centering
    \begin{subfigure}{.32\textwidth}
    \includegraphics[width=\textwidth]{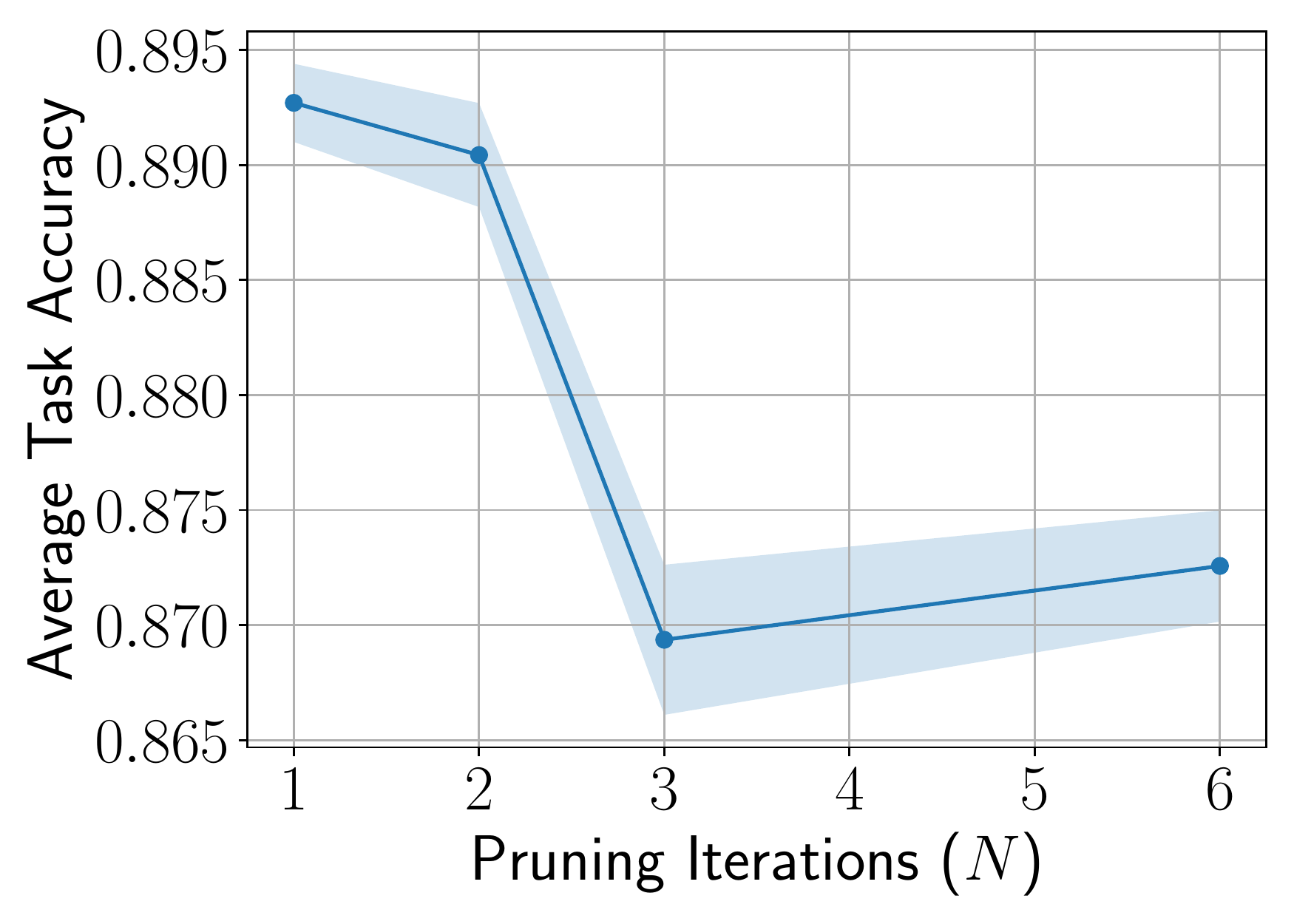}
    \caption{PackNet: 0.8 Pruning}
    \label{fig:packnet08cifar10}
    \end{subfigure}
    \begin{subfigure}{.32\textwidth}
    \includegraphics[width=\textwidth]{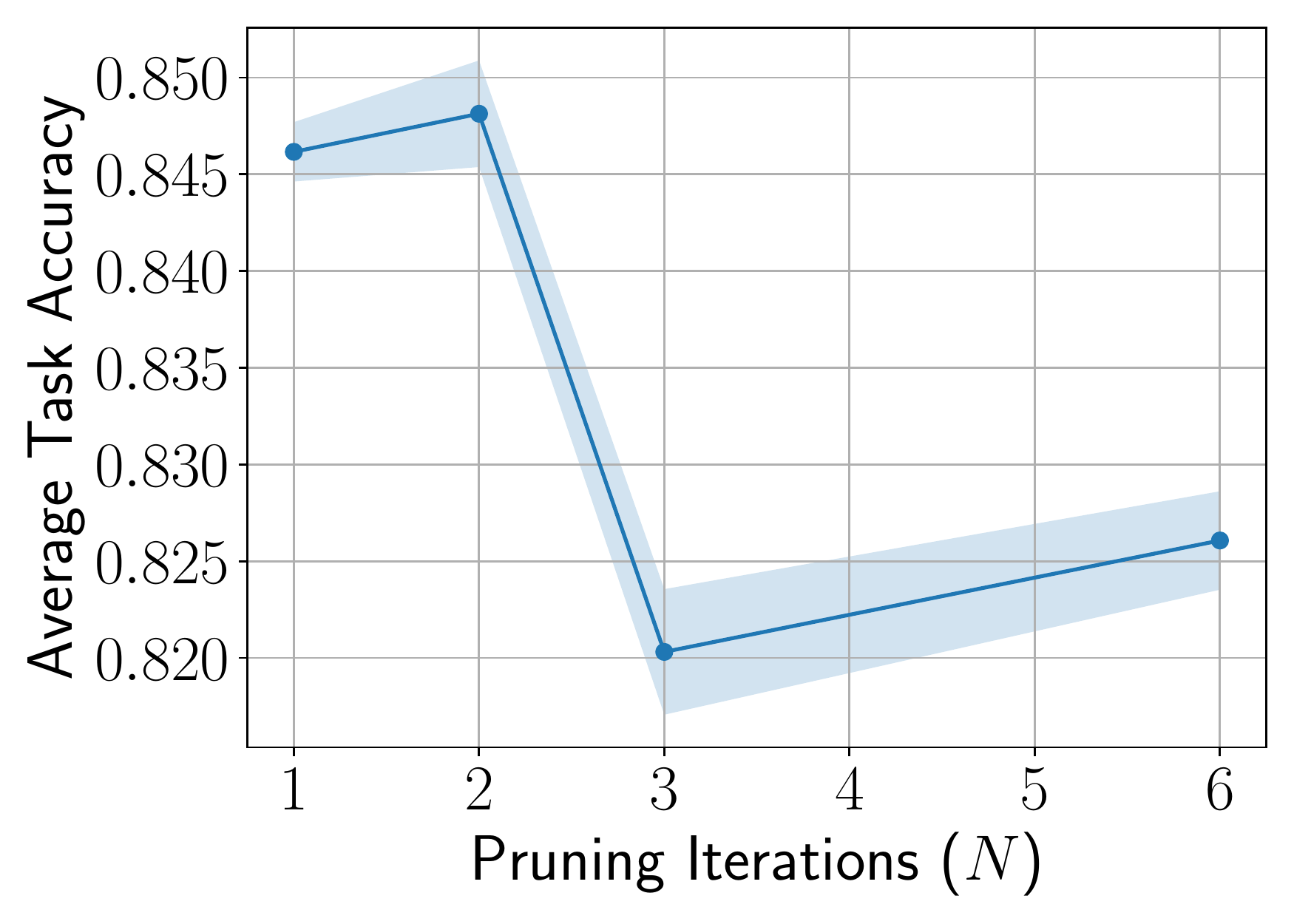}
    \caption{PackNet: 0.9 Pruning}
    \label{fig:packnet09cifar10}
    \end{subfigure}
    \begin{subfigure}{.32\textwidth}
    \includegraphics[width=\textwidth]{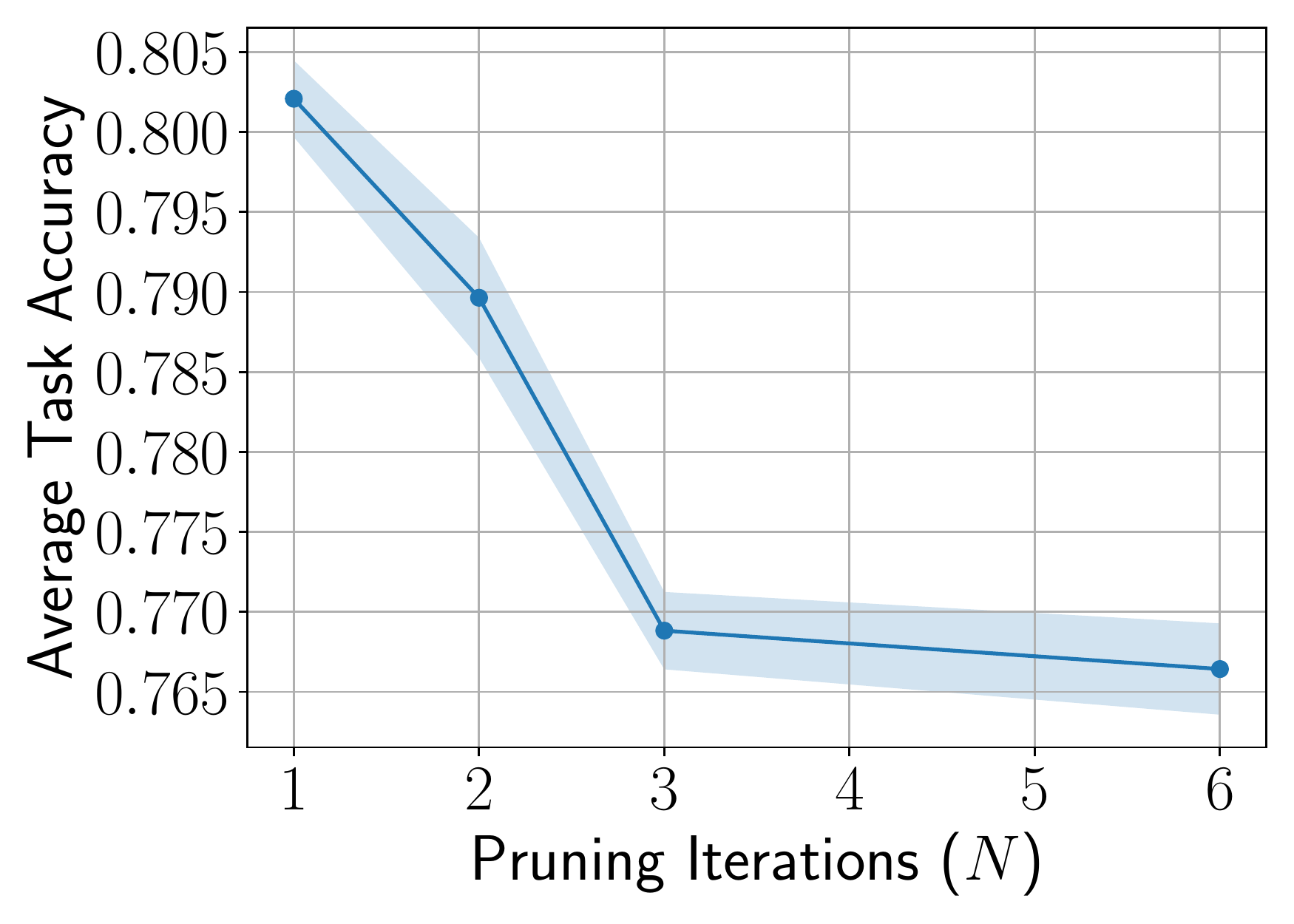}
    \caption{PackNet: 0.95 Pruning}
    \label{fig:packnet095cifar10}
    \end{subfigure}
    \caption{Iterative pruning results by fixing the time budget and varying the number of pruning rounds (mean \& standard error from 10 runs). \textbf{Above}: MNIST. \textbf{Below}: CIFAR10.}
    \label{fig:pruningexpall}
\end{figure}

We utilize the same datasets from Section \ref{sec:weightefficiencyresults}, and constrain the retraining time budget $B$ for both MNIST and CIFAR10 networks to 72 and 6 epochs respectively, presenting these results in Figure \ref{fig:pruningexpall}.
We see that it is vital to select the correct number of iterations to prune over, especially at higher levels of pruning. For instance, in MNIST, the difference between the best and worst performing networks at $99.5\%$ pruning causes a near doubling in the error rate (from $1.5\%$ to $3\%$), and we observe that the best choice of $N$ at $99.5\%$ pruning performs nearly as well as the worst at $99\%$ pruning, meaning that for the same network we achieve a further $2\times$ compression through careful selection of pruning iterations. Similarly in CIFAR10, at $95\%$ compression, there is a $3.5\%$ difference between the best and worst choices of $N$. For comparison, in Appendix \ref{app:nobudget} we validate the existing literature that more pruning iterations gives better performance only if there is no fixed time budget $B$.

Moreover, these two network architectures have radically different behavior when considering optimal pruning iteration $N$. 
In CIFAR10 lower pruning iterations, oftentimes one-shot, yield the best performance, while the opposite is true for MNIST. We note however an interesting trend in the MLP networks; at some point performance drops when increasing iterative pruning iteration, but then improves beyond the best performance so far. It may be the case that we are in this `performance drop' region with the CIFAR10 experiments; this merits further investigation into understanding why this regime exists, what it represents, and whether there is a biological connection.
\section{Discussion}
In this paper we have analyzed and extended an existing popular continual learning approach using insights from human learning. This framing has allowed us to determine that methods using weight freezing can cause a reduction in the memory efficiency of neural networks, sometimes requiring over $2\times$ as many weights to achieve the same level of performance. We have also made analogies between iterative pruning and sleeping in humans, specifically the similarities of sleep cycles and iterative pruning. With this in mind, we have applied the constraint of a time budget to the iterative pruning comparisons, and note that more pruning iterations does not necessarily translate into better performance. Instead the optimal number of iterations appears to be dependent on network architecture, and a poor choice can diminish performance significantly.

For future work we will extend the empirical analysis to more network architectures and additional data sets. We would also investigate new strategies of iterative pruning by taking inspirations from biological learning. For example, we would consider the setting where pruning itself has an associated time-cost, and consider non-uniform epoch allocation for different pruning iteration, 
similar to how sleep cycles change in length through the night \cite{sleepcycle2007}.

\bibliography{refs}

\appendix
\section{Implementation Details}
\label{app:imp}

For the MLP and VGG experiments we use the following hyperparameters in Table \ref{tab:hyperparam}. These were found to deliver the best validation performance using the default PackNet implementation and one-shot pruning.
\begin{table}[htb]
    \centering
    \small
    \begin{tabular}{@{}ccc@{}}
    \toprule
    \multirow{2}*{\vspace{-0.1cm} Hyperparameter} & \multicolumn{2}{c}{Network Type} \\ \cmidrule{2-3}
     & MLP & Convolutional  \\ \midrule
        Optimizer & \textit{Adam} & \textit{Adam} \\
        Learning Rate & \num{1e-3} & \num{1e-4} \\
        Batch Size & 128 & 32 \\
        Training Epochs ($E$) & 72 & 12 \\
        Retraining Budget ($B$) & 72 & 6 \\
        Activation Function & ReLU & ReLU \\
    \bottomrule
    \end{tabular}
    \caption{Hyperparameters used in training}
    \label{tab:hyperparam}
\end{table}

The fully connected architecture used for the MLP network is $784:256:256:2$, and for the CIFAR10 network is conv3-32, conv3-64, maxpool, conv3-128, conv3-128, maxpool, dropout(0.1), fc8192, dropout(0.1), fc1024, fc2.

\section{Matching Experiment}
\label{app:match}

\begin{figure}[htb]
    \begin{subfigure}{.33\textwidth}
    \includegraphics[width=\textwidth]{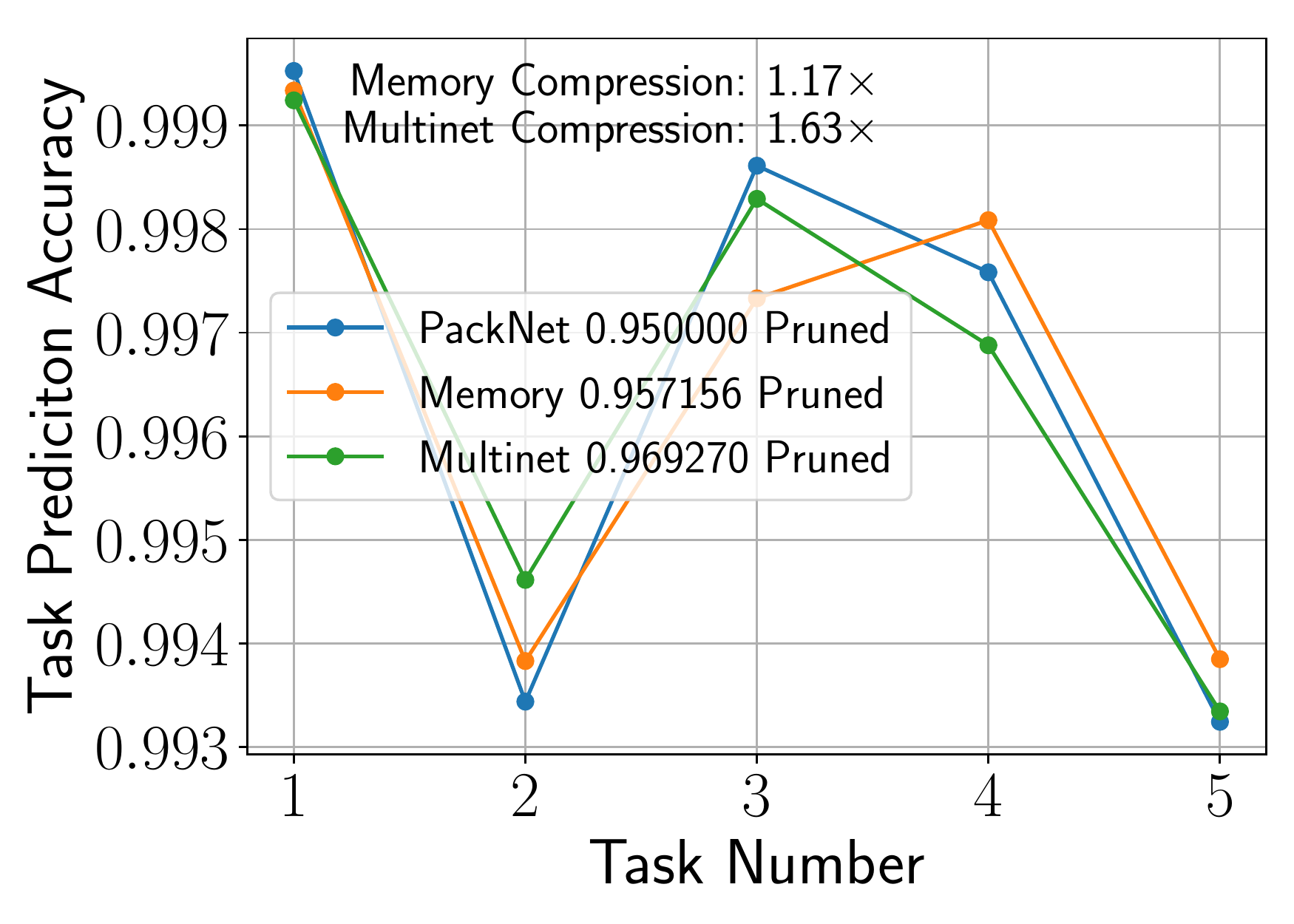}
    \caption{PackNet 0.95 Pruning, Ave. Perf.: 0.996}
    \label{fig:compare_095_mnist}
    \end{subfigure}%
    \begin{subfigure}{.33\textwidth}
    \includegraphics[width=\textwidth]{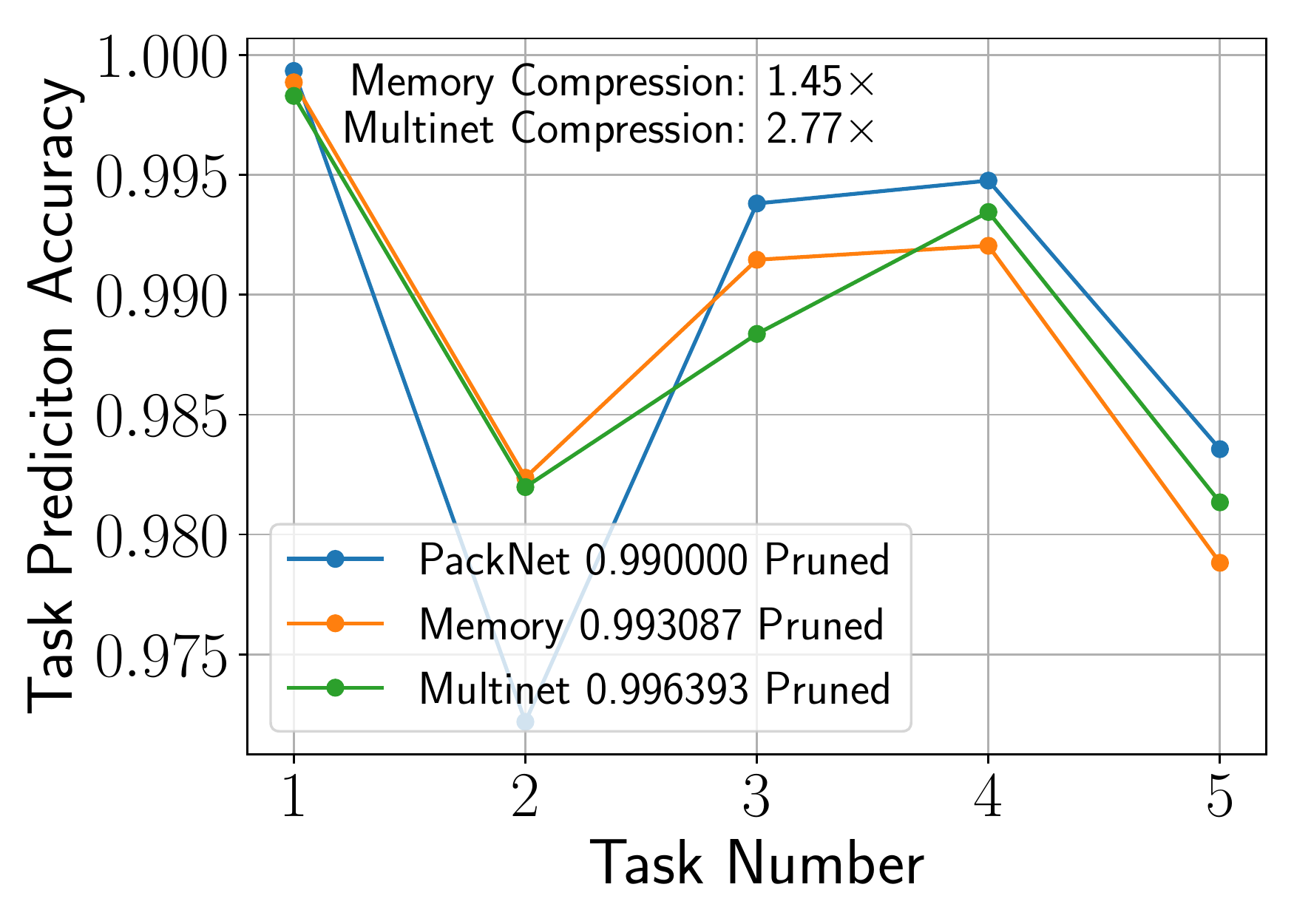}
    \caption{PackNet 0.99 Pruning, Ave. Perf.: 0.989}
    \label{fig:compare_099_mnist}
    \end{subfigure}%
    \begin{subfigure}{.33\textwidth}
    \includegraphics[width=\textwidth]{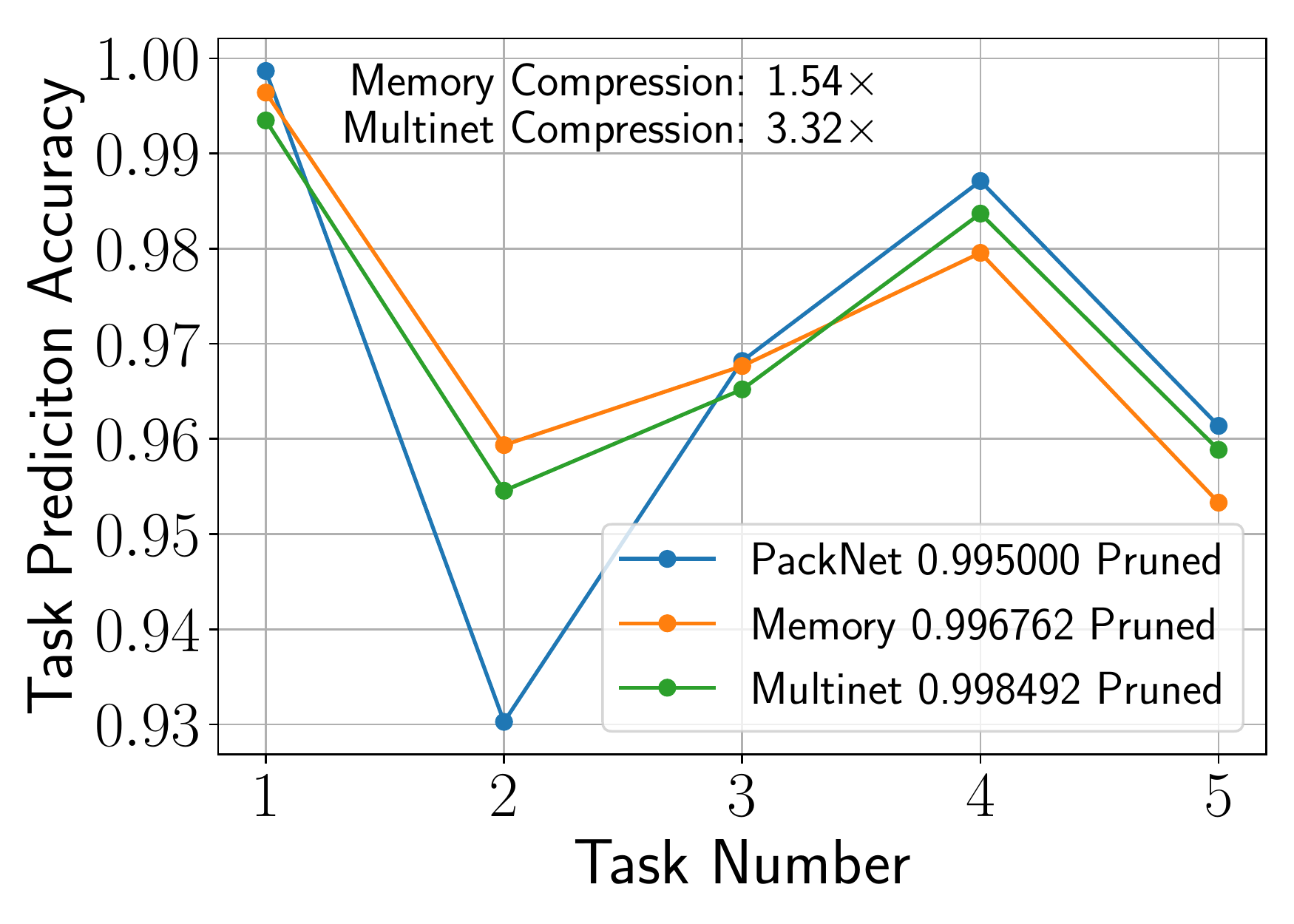}
    \caption{PackNet 0.995 Pruning, Ave. Perf.: 0.970}
    \label{fig:compare_0995_mnist}
    \end{subfigure}
\\
    \begin{subfigure}{.33\textwidth}
    \includegraphics[width=\textwidth]{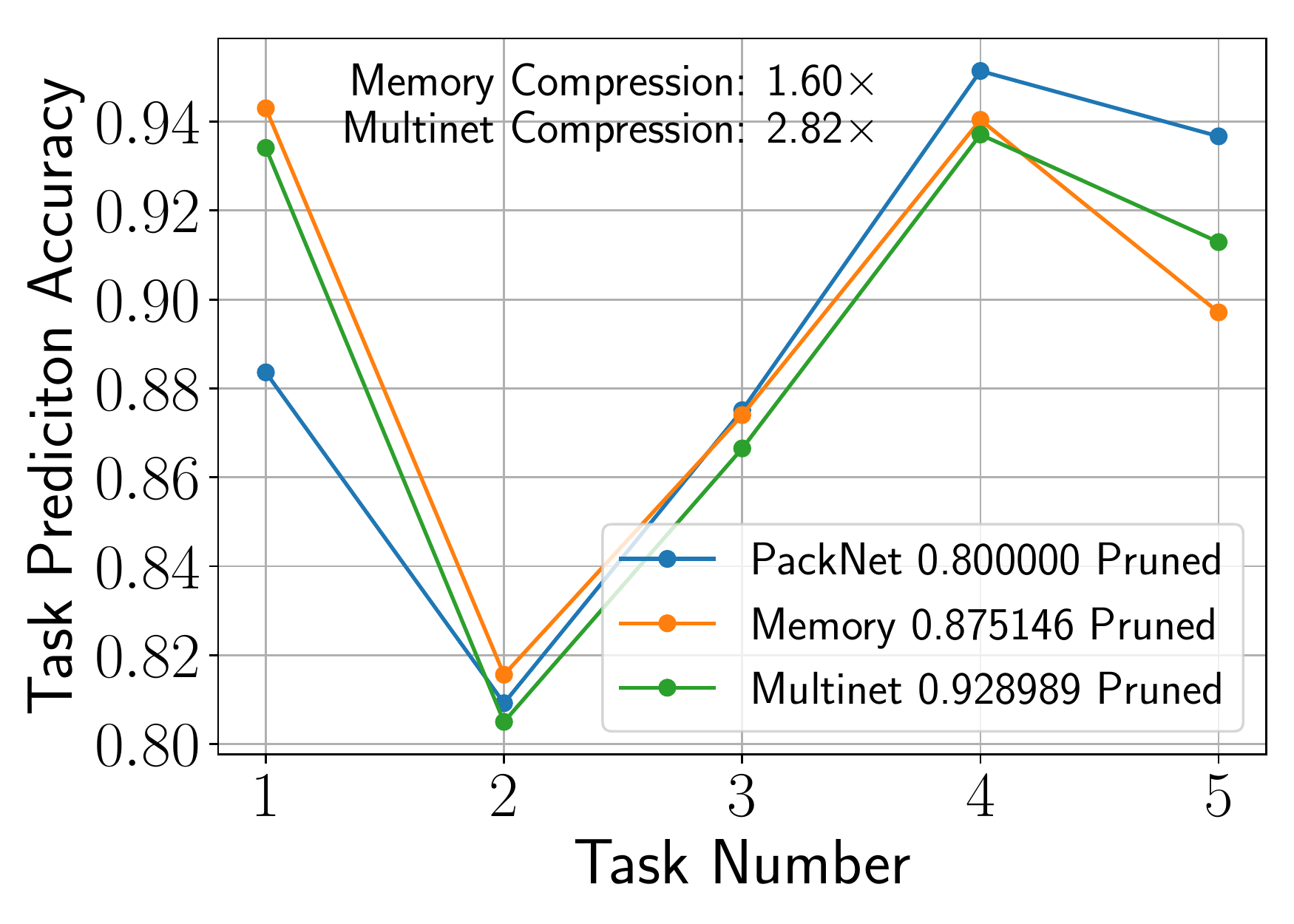}
    \caption{PackNet 0.8 Pruning, Ave. Perf.: 0.891}
    \label{fig:compare_08_cifar10}
    \end{subfigure}%
    \begin{subfigure}{.33\textwidth}
    \includegraphics[width=\textwidth]{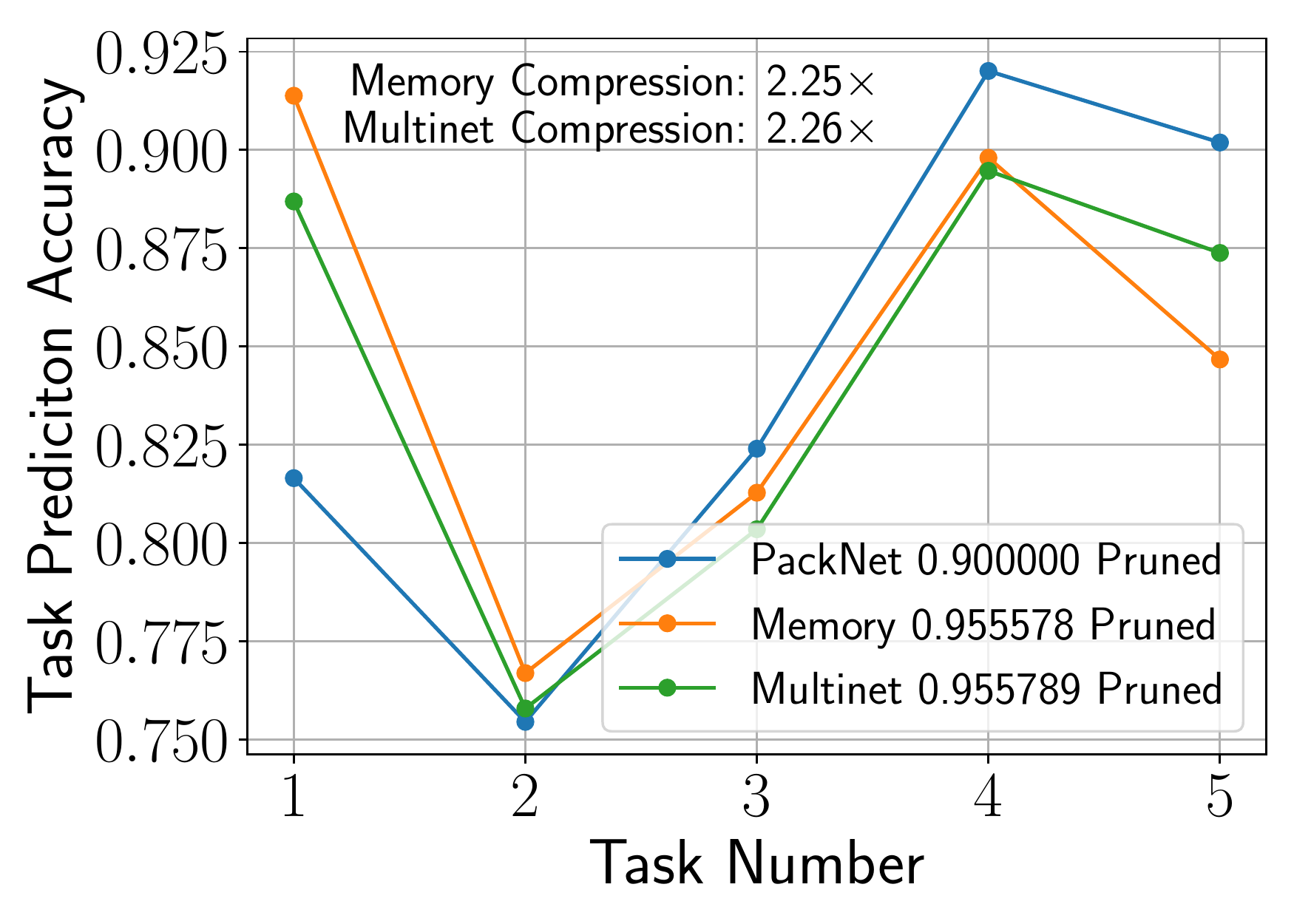}
    \caption{PackNet 0.9 Pruning, Ave. Perf.: 0.843}
    \label{fig:compare_09_cifar10}
    \end{subfigure}%
    \begin{subfigure}{.33\textwidth}
    \includegraphics[width=\textwidth]{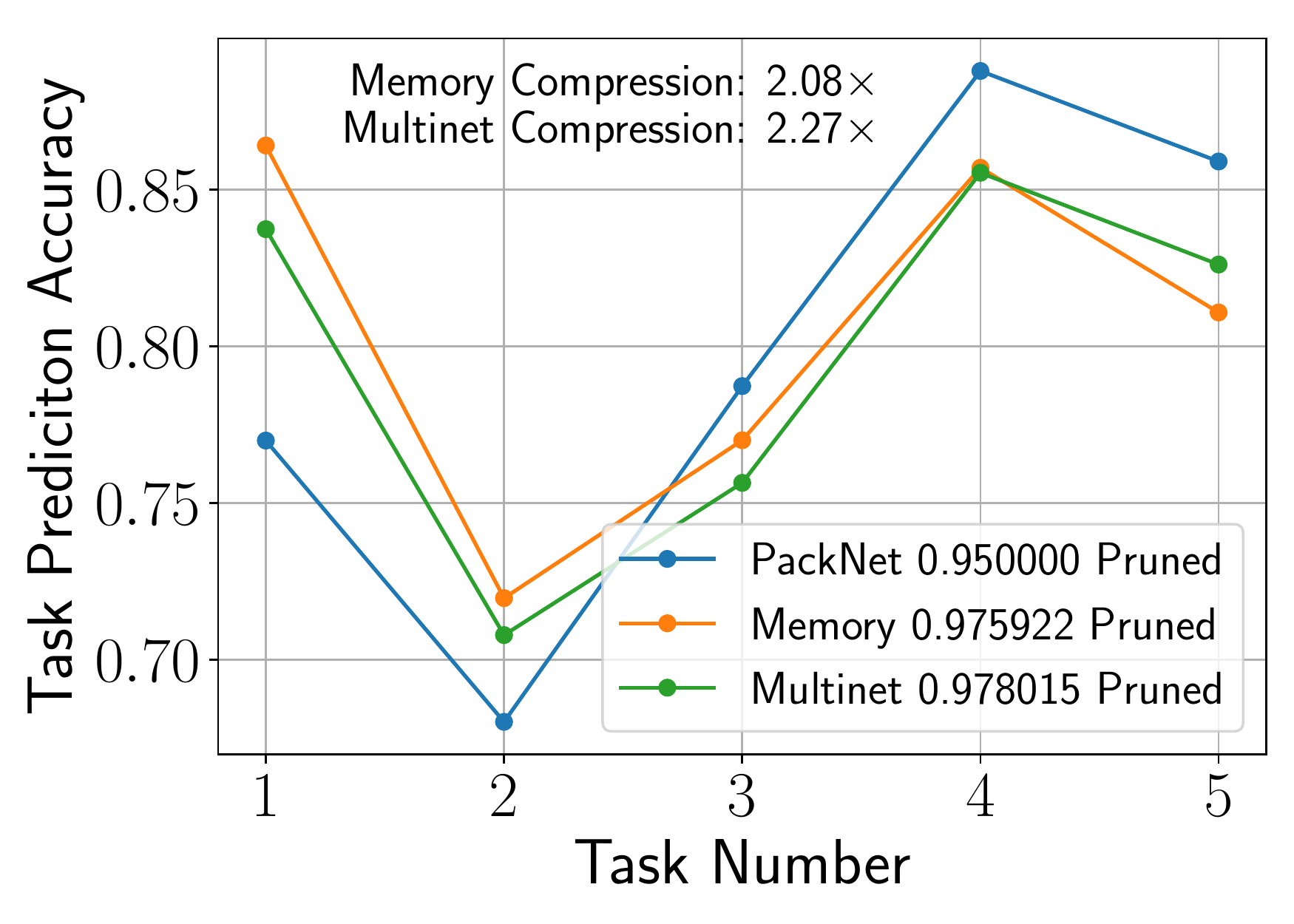}
    \caption{PackNet 0.95 Pruning, Ave. Perf.: 0.797}
    \label{fig:compare_095_cifar10}
    \end{subfigure}
    \caption{Matching Results per Task. \textbf{Above}: MNIST. \textbf{Below}: CIFAR10.}
    \label{fig:match_details}
\end{figure}

In order to match the performance of each idealized case (i.e., Memory and Multinet) to the performance of Packnet, we first train PackNet at the desired $p$ (i.e., 0.8) for 5 seeds, and determine the average performance over all tasks and seeds; this is called $\text{PerfPN}_p$. We then define the performance Multinet and Memory approaches over 5 seeds at a pruning $\bar{p}$ as $\text{PerfMN}_{\bar{p}}$ and $\text{PerfMem}_{\bar{p}}$ respectively. We now aim to find a values of $\bar{p}$ such that $\text{PerfPN}_p \approx \text{PerfMN}_{\bar{p}}$ and $\text{PerfPN}_p \approx \text{PerfMem}_{\bar{p}}$. To do this we use a scalar minimization algorithm in SciPy ($\text{minimize\_scalar}$) that finds values of $\bar{p}$ which ensure approximately equal performance. We find in practice minimizing the square of the performance differences gives more accurate matching (i.e., minimizing $(\text{PerfPN}_p - \text{PerfMN}_{\bar{p}})^2$). We present the per-task breakdown of the matching experiment in Figure \ref{fig:match_details}.

\section{Iterative Pruning without a Budget}

\begin{figure}[t]
    \centering
    \begin{subfigure}{.32\textwidth}
    \includegraphics[width=\textwidth]{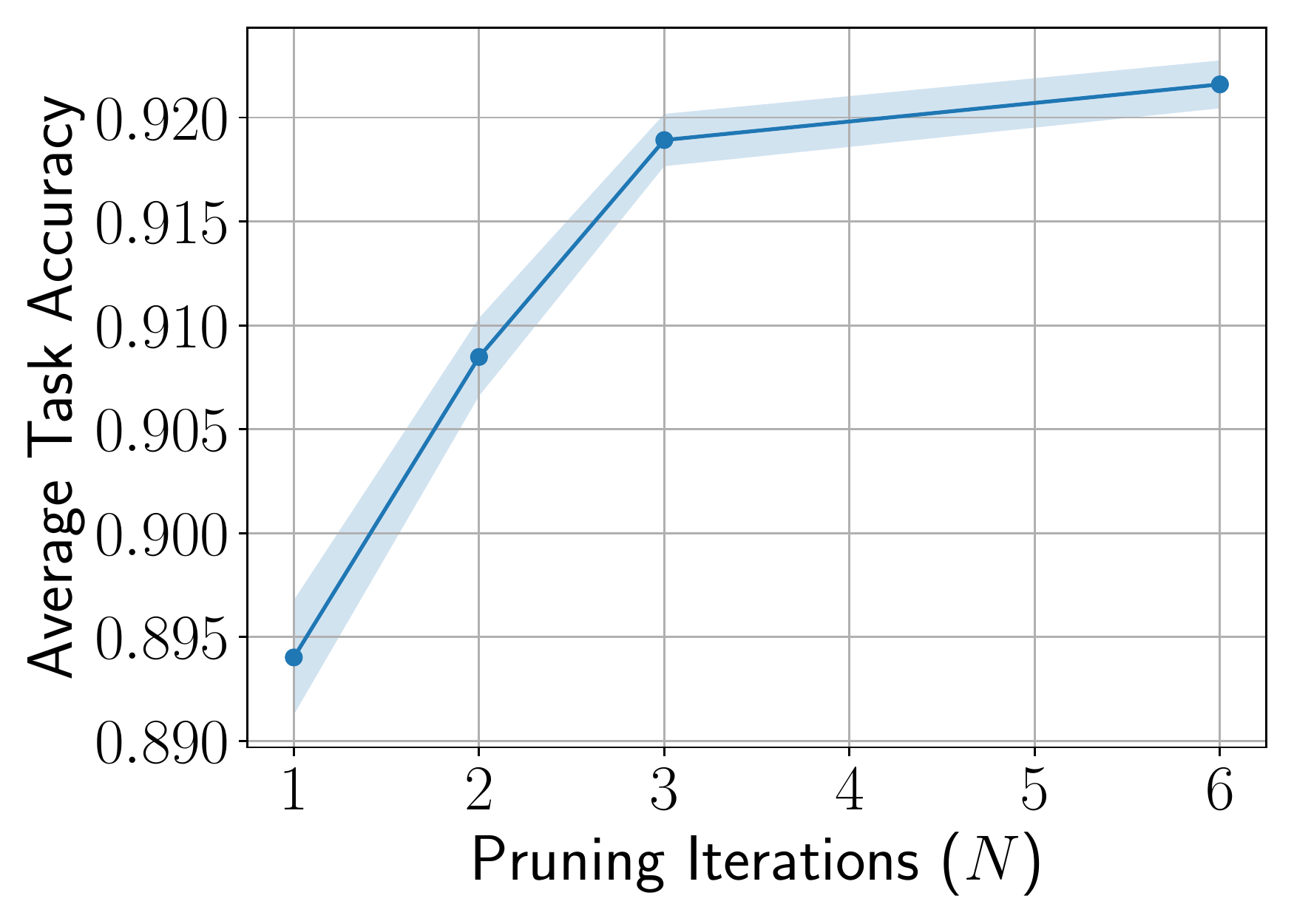}
    \caption{PackNet: 0.8 Pruning}
    \label{fig:packnet08cifar10nobudget}
    \end{subfigure}
    \begin{subfigure}{.32\textwidth}
    \includegraphics[width=\textwidth]{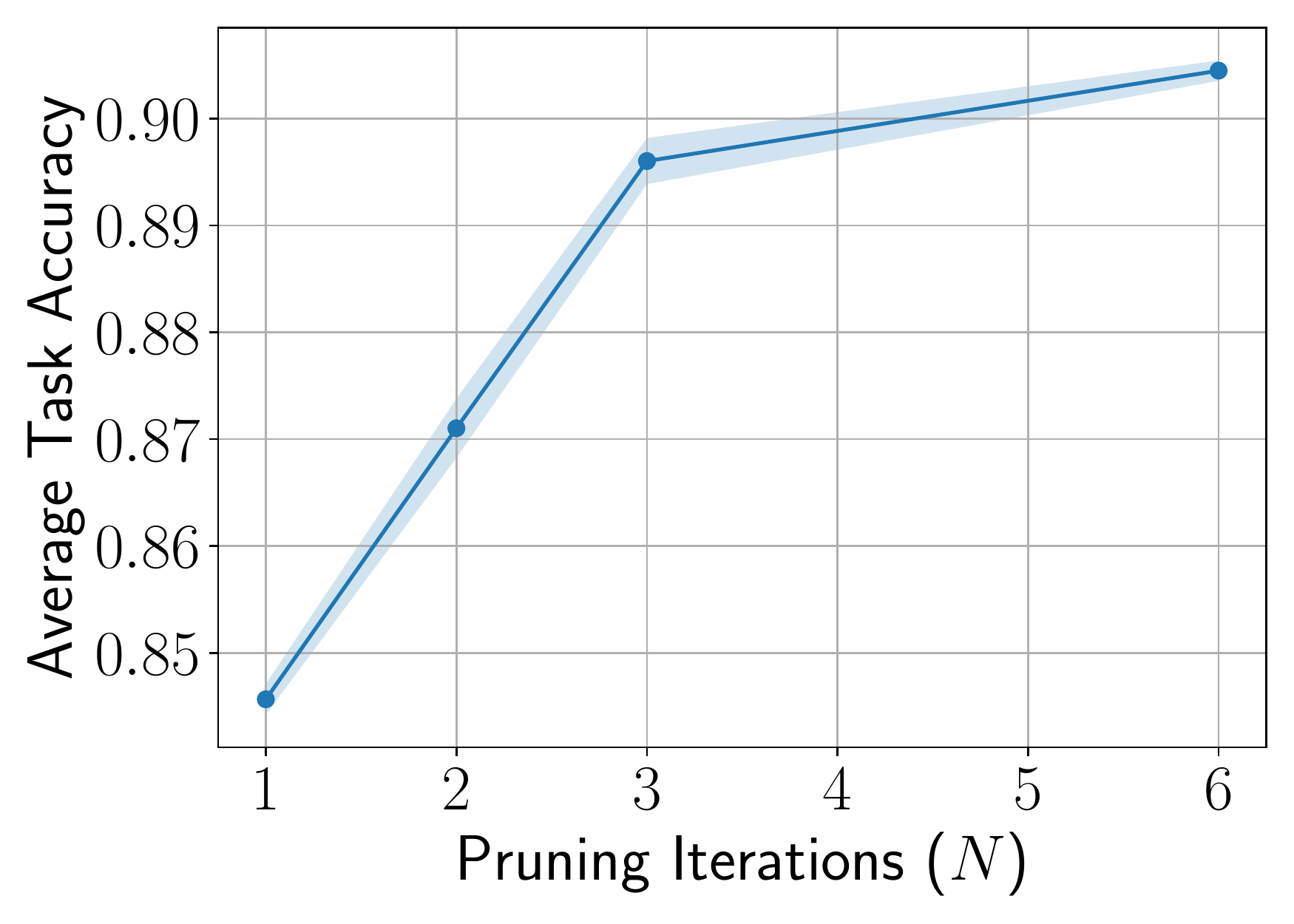}
    \caption{PackNet: 0.9 Pruning}
    \label{fig:packnet09cifar10nobudget}
    \end{subfigure}
    \begin{subfigure}{.32\textwidth}
    \includegraphics[width=\textwidth]{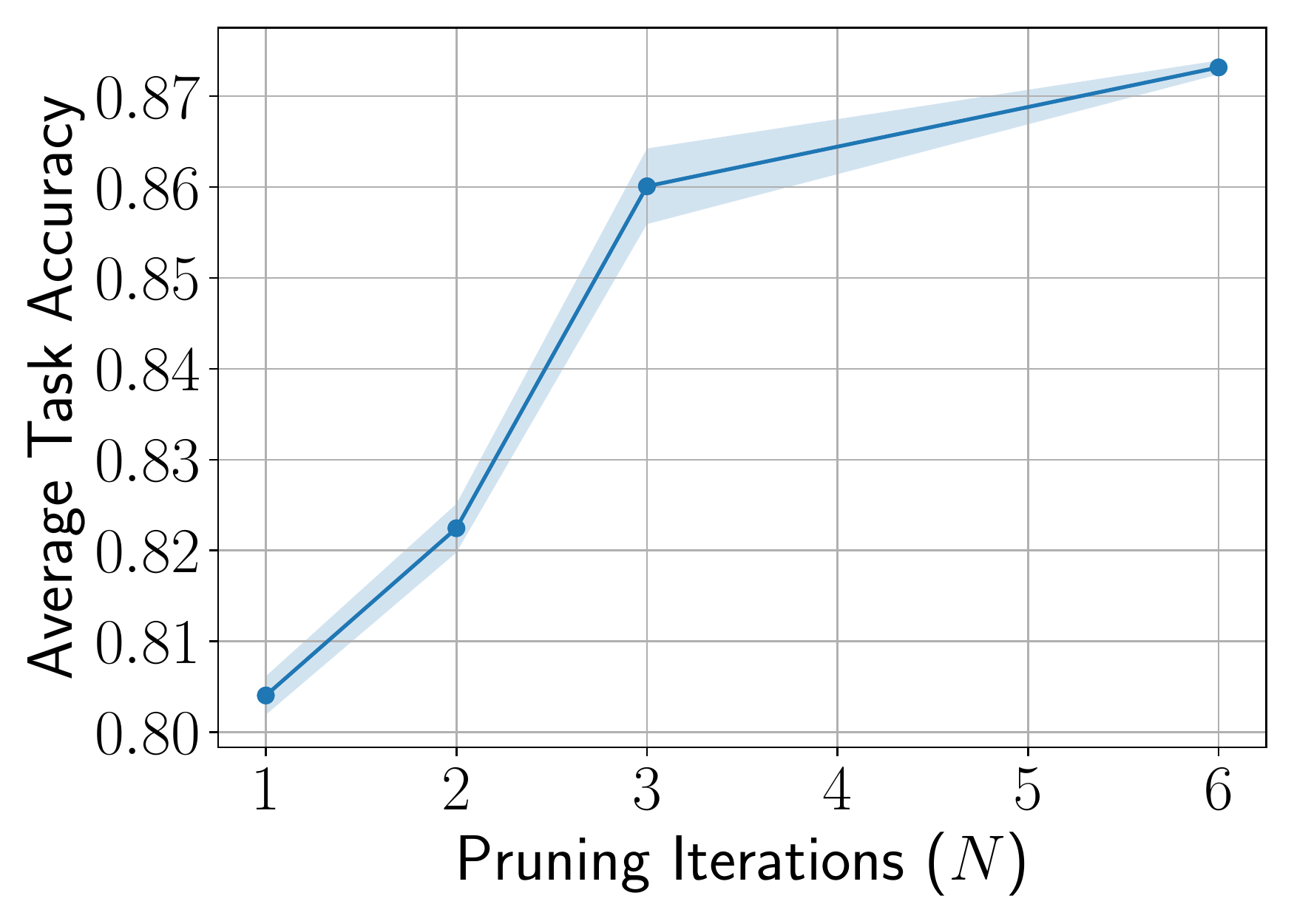}
    \caption{PackNet: 0.95 Pruning}
    \label{fig:packnet095cifar10nobudget}
    \end{subfigure}
    \caption{Iterative Pruning No Budget Results for CIFAR10.}
    \label{fig:pruningexpallnobudget}
\end{figure}

As observed in \cite{iterative2015}, using iterative pruning to achieve the desired level of pruning instead of one-shot pruning can greatly increase performance, but only if each round of iterative pruning uses as many epochs as the one-shot approach. We show this result is true for illustrative purposes in Figure \ref{fig:pruningexpallnobudget}. However because this approach scales as $\mathcal{O}(N)$, this is not practicable.

\label{app:nobudget}

\section{Algorithm to Determine Per Iteration Pruning}

In Algorithm \ref{alg:getprunepct} we present how to calculate the amount of pruning per round that we used in our experiments.

\begin{algorithm}
\SetAlgoLined
\SetKwInOut{Input}{Input}
 \Input{Number of iterative pruning rounds $N$, Number of tasks $|T|$, Final prune proportion $p$}
 $r = \exp(\log(p)/|T|)$\;
 $z = 1 - \exp(\log(1 - r)/N)$\;
 return $z$\;
 \caption{Calculate Pruning Percentage $z$}
 \label{alg:getprunepct}
\end{algorithm}

\end{document}